\documentclass[letterpaper]{article}
\usepackage{aaai}
\usepackage{times}
\usepackage{helvet}
\usepackage{courier}
\usepackage{graphicx}
\usepackage{amsmath}
\usepackage{booktabs}
\usepackage{url}
\usepackage{hyperref}
\usepackage{xcolor}
\begin{document}
% The file aaai.sty is the style file for AAAI Press 
% proceedings, working notes, and technical reports.
%
\title{Exploring Genre and Success Classification through Song Lyrics using DistilBERT: A Fun NLP Venture}
\author{Servando Pizarro, Moritz Zimmermann, Miguel Serkan Offermann, Florian Reither
\\
University of Cologne\\
Albertus-Magnus-Platz\\
50923 Cologne\\
}

\maketitle
\begin{abstract}
\begin{quote}
This paper presents a natural language processing (NLP) approach to the problem of thoroughly comprehending song lyrics, with particular attention on genre classification, view-based success prediction, and approximate release year. Our tests provide promising results with 65\% accuracy in genre classification and 79\% accuracy in success prediction, leveraging a DistilBERT model for genre classification and BERT embeddings for release year prediction. Support Vector Machines outperformed other models in predicting the release year, achieving the lowest root mean squared error (RMSE) of 14.18. Our study offers insights that have the potential to revolutionize our relationship with music by addressing the shortcomings of current approaches in properly understanding the emotional intricacies of song lyrics.
\end{quote}
\end{abstract}

\section{Introduction}
In the modern world, music is essential to our existence since it affects our feelings, entertains us, and fosters cross-cultural relationships. Additionally, categorizing songs by genre has also practical implications for recommendation systems and personalized content delivery. But there's a big problem when people try to understand a song's emotional undertones just by reading the words. When the melody is absent and one reads the lyrics alone, the spirit and mood of the music are frequently lost, and it is hard to interpret the emotional tone of the lyrics. People are unable to completely appreciate and comprehend the depth of musical expression as a result of this constraint. That’s why we want to figure out if it is possible to predict essential information about a song, including its genre, approximate age, and even its potential success in terms of page views. Addressing this problem, we want to use natural language processing (NLP) techniques to analyze and interpret song lyrics comprehensively. We want to gain important insights into the world of music through the suggested solution, increasing everyone's enjoyment and accessibility. This project has the potential to completely transform our understanding of and relationship to music by overcoming the constraints imposed by the sole focus on melody. 

While addressing the challenge of song lyrics classification, we draw insights from existing works such as Khrulchenko (2020) and \cite{rajendran2022lybert}. On the one hand, Khrulchenko (2020) employed traditional machine learning methods, while on the other hand, \cite{rajendran2022lybert} explored NLP-based models. Even though these works provide foundational knowledge, limitations persist. We aim to build upon these approaches and add value to the world of music genre prediction. 

This paper aims to determine which model is the most effective and promising for classifying song genres, the success of songs and for predicting the release year. For the classification problem, we use a DistilBERT model (a distilled version of Bidirectional Encoder Representations from Transformers). To encode the song lyrics, a DistilBERT tokenizer was utilized. For our dependent variables, we split the 'genre' and 'views' into five genre classes and two success classes. Then we use the created lyrics tokens -  based on the pre-trained DistilBERT model 'distilbert-base-uncased' - to predict the four genres and if the songs were a success or not. The task of predicting musical genres presented significant challenges, particularly in distinguishing between pop and rock lyrics. Additionally, forecasting the success by views was a challenge because there is a high range of views, and it was difficult to find the right threshold value for a song's success. 
For predicting the release year of songs, we used BERT embeddings and machine learning models. The process commenced with the extraction of BERT embeddings from song lyrics, incorporating the DistilBERT tokenizer and model. Based on the embeddings and the dependent variable 'year', we figured out the best predictive machine learning model with the lowest root mean squared error (RMSE). 

The paper is structured as follows: The next section provides information about related research that already exists. Then we go deeper into our methodology where we explain which methods we use for our song classification problem. After that, we offer more insights about the Genius Dataset and the results of our trained models. Finally, the last section offers concluding remarks.   

\section{Related Work}
In the following section, we will briefly analyze related work. There is not much literature out there attempting to address the problem of classifying song lyrics to their respective genre. However, most researchers implemented classical approaches like machine learning models or simple neural networks to do a song lyrics classification.

\subsection{Application's perspective}
From the application’s perspective, Khrulchenko (2020) has been one key contributor to song lyrics classification. In his approach, ‘Pop, Rap and Heavy Metal - lyrics classifier’, Khrulchenko attempted to classify the song genres using traditional machine learning techniques such as Logistic regression, linear SVC, decision tree, and gradient descent. However, the issue is that traditional methods often face limitations in capturing the complex patterns and semantics of music lyrics. As a result, music genre prediction oftentimes becomes imprecise. Within our NLP venture, we aim to address the limitations and challenges of conventional machine learning approaches, such as  Khrulenko (2020), by leveraging state-of-the-art natural language processing models, particularly DistilBERT. By using DistilBERT’s contextual understanding, our approach aims to improve the accuracy and reliability of genre \& success classification.

\subsection{Methodology's perspective}
From a methodology perspective, sequence models have been used in quite different tasks, such as prediction \cite{li2020tensor,tsung2020discussion,li2020long,mao2022jointly,lin2023dynamic,wang2023correlated,jiang2023unified,chen2023adaptive,ruan2023privacy,xu2023kits,li2023critical,li2024non,liu2024spatial,guo2024online,sergin2024low,ye2024survey,liu2024timecma,zhang2024dualtime}, completion \cite{li2023visiontraj}, classification \cite{ziyue2021tensor,li2022individualized,liu2023relation,li2023tensorb,li2023choose,li2023tensor}, anomaly detection \cite{li2022profile,yan2024sparse}, agent-based decision making \cite{mao2024pdit,lu2024dualight,mao2022transformer,jiang2024x,ruan2024coslight}, LLMs \cite{ruan2023tptu,kong2024tptu,zhang2024controlling}, and so on. Specifically, we drew inspiration from \cite{rajendran2022lybert}. Rajendran et al. (2022) also employed NLP models for classifying song lyrics. They implemented a LyBERT-based model to predict the emotional tone of a song's lyrics. To do so, they divided the lyrics into four categories: happy, relaxed, angry, and sad. In contrast to his work, we go a step further by incorporating other features such as genre, success, and release year just by looking at the song lyrics. These improvements not only help us to understand song lyrics better, but they also allow us to gain deeper understanding and connections within the music domain.

Our goal is to add value to the existing literature by refining and extending the approaches based on Khrulenko (2020) and \cite{rajendran2022lybert}. By capitalizing on NLP approaches and adding new features, we hope to overcome the limitations of traditional methods and enhance the analysis of song texts. As a result, we hope to advance the area of song lyrics' predictive analysis.

\section{Methodology}
In this paper, we use a version of BERT (Bidirectional Encoder Representations from Transformers), which is an important natural large language processing (NLP) model introduced by Google in 2018. BERT uses a bidirectional transformer design, which enables it to analyze both preceding and subsequent words simultaneously, allowing it to examine a word's whole context, in contrast to typical language models that process text sequentially. Large volumes of unlabeled text are exposed to the model during its pre-training phase, which helps it acquire complex representations of language. 
For classifying genre and success based on song lyrics, we use DistilBERT which is a transformer model that is compact, lightweight, quick, inexpensive, and trained by distilling BERT basis. The pre-trained distilbert-base-uncased model is enabling us to figure out the complex relationships embedded within songs. This section is based on research of \cite{adel2022improving}.

\subsection{Distilled BERT for Feature Extraction}

In the following, we go deeper into the methodology and architecture of DistilBERT. 
The suggested feature extraction model, which is based on DistilBERT, goes through several stages to process input sequences, which are tweets.
\begin{figure}[h]
\centering
\includegraphics[width=\linewidth]{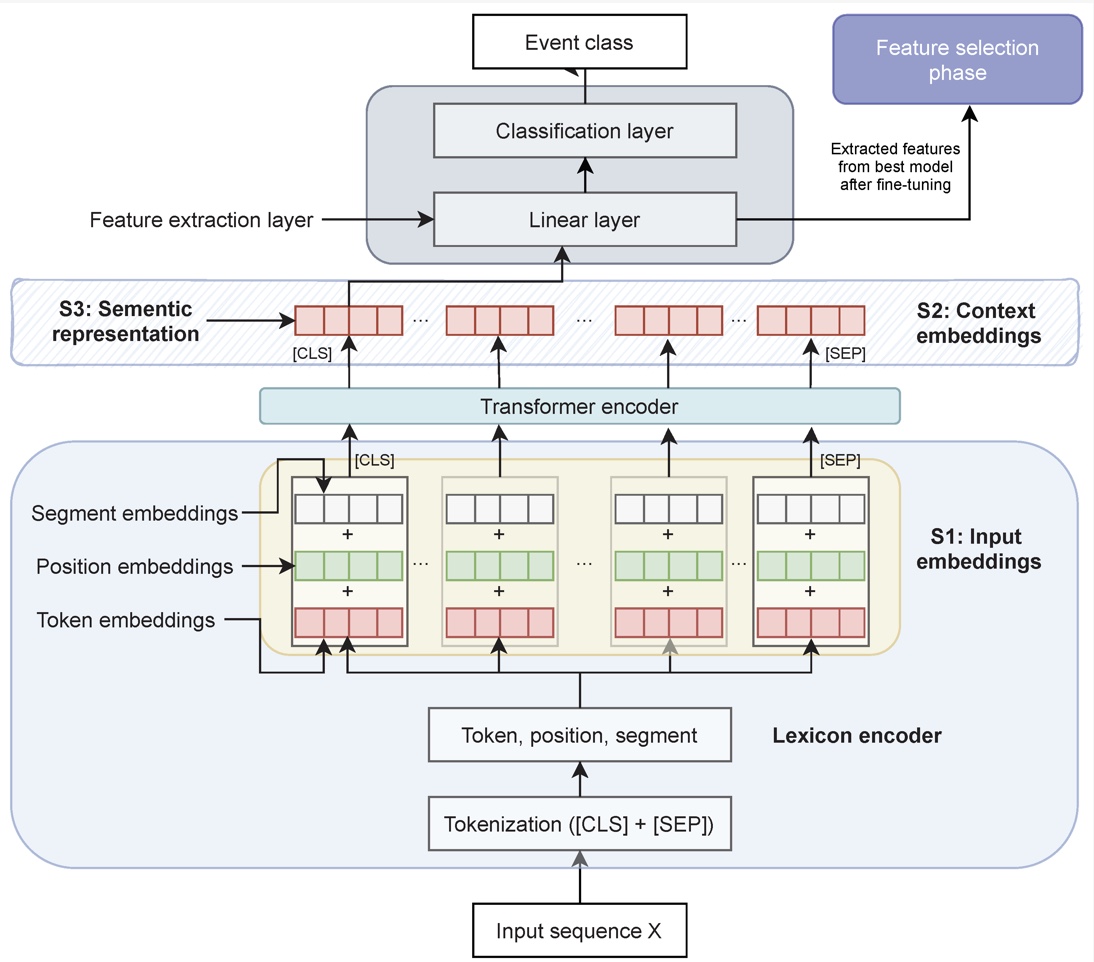}
\caption{The proposed feature extraction model}
\end{figure}
The architecture (shown in Figure 1) turns input sequences (X) into embedding vectors (S1) for individual words using DistilBERT. Utilizing a self-attention mechanism, DistilBERT's transformer encoder creates contextual embeddings (S2) that encompass each word's contextual information. The entire tweet's semantic content is then represented by a single vector (S3) that is created by concatenating these contextual embeddings. A fully connected layer receives the concatenated vector (S3) as input, and outputs an output vector of size 'd,' where 'd' is the number of neurons. Then, in order to optimize the previously trained DistilBERT for the particular purpose of genre \& success detection, a classification layer is used. The purpose of this classification layer is to forecast the genre \& success class linked to every tweet sequence that is input. 

\subsection{Lexicon Encoder}
A multi-layered Neural Network is used to process each tweet, which is represented by a series of tokens with a length of s. Tokens make up the input $X = x_1,..., x_s$. Each token is mapped to an associated embedding vector that represents word, segment, and positional information. A special token [CLS] is inserted as the first token ($x_1$) in accordance with the encoding strategy suggested by \cite{devlin2018bert}, and the [SEP] token is positioned at the end of the sequence. The embedding vectors for $X$ are produced by the lexical encoder. For every token in the sequence, this entails adding up the word, segment, and positional embeddings. 

\subsection{Transformer Encoder}

Now, we focus more on the general DistilBERT model architecture, which is shown in Figure 2. We use a pre-trained multilayer bidirectional transformer encoder to convert input vectors (S1) into contextual embedding vectors. 
\begin{figure}[h]
\centering
\includegraphics[width=\linewidth]{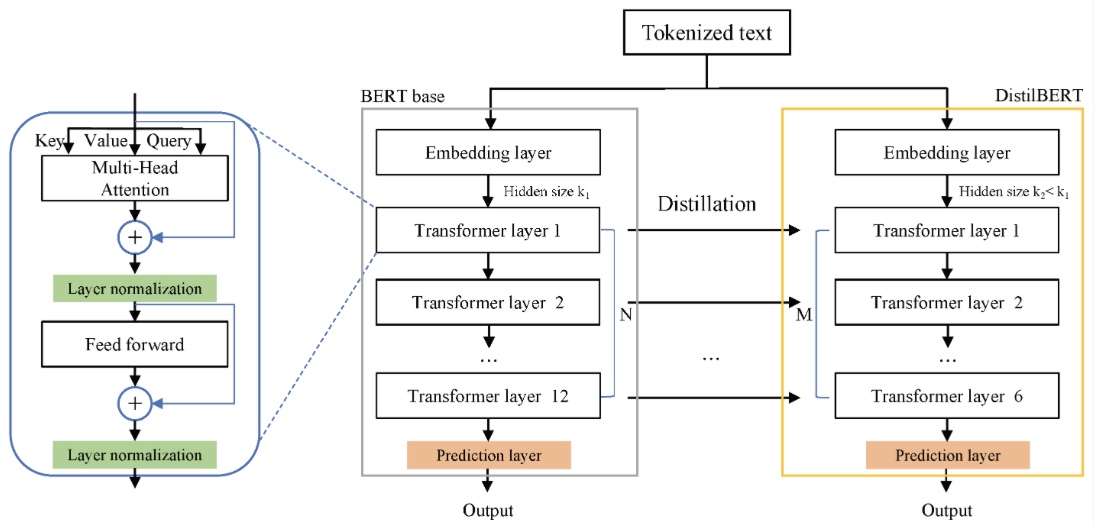}
\caption{The DistilBERT model architecture and components}
\end{figure}
DistilBERT employs knowledge distillation to reduce the BERT base model (bert-base-uncased) parameters. It runs 60 \% faster and requires 40 \% less parameters than bert-base-uncased while maintaining over 95 \% of BERT's performance \cite{sanh2019distilbert}. With the use of distillation, one can approximate the whole output distributions of BERT by utilizing a more compact model called DistilBERT, which has six transformer layers instead of 12. With 66 million trainable parameters, DistilBERT performs noticeably better than the 110 million parameter original BERT base model. Similar to the BERT base training data, the 16 GB of training data for DistilBERT comes from the English Wikipedia and the Toronto books corpus. Gradient accumulation and a batch size of 16 are used during DistilBERT's training. To increase efficiency, this method aggregates gradients from several mini-batches before modifying the parameters. Notably, the training approach omits next sentence prediction (NSP) and segment embeddings learning objectives. When combined, these modifications simplify the training process and make DistilBERT a viable and productive substitute for BERT in a range of natural language processing applications.

\subsection{Fine-Tuning on Genre and Success Classification Task}
In our training loop, we initialize an AdamW optimizer with a learning rate of 1e-5 and a weight decay term of 0.01 which introduces L2 regularization helping to prevent overfitting. Additionally, we use the CrossEntropyLoss to compute the loss between predicted logits and ground truth labels during training.

The final aim is to formulate a multi-class classification issue with the contextual embedding learnt by the [CLS] token from the input tweet $X$, or semantic representation S3. Predicting the likelihood that c will fall into a particular class, 'c,' which stands for a genre or the success, is the aim.
\begin{equation}
P(c|X) = \text{Softmax}(W^T \cdot X)
\end{equation}
Equation (1) specifies the use of the Softmax function to determine this categorization probability. The Softmax function makes it easier to convert raw scores into probability distributions, which allows the model to forecast the chance that a given genre or success will include $X$.

\subsection{Prediction of the Release Year based on BERT Embeddings}
For our regression model task, we extract BERT embeddings to predict the release year of songs. We used again a pre-trained DistilBERT tokenizer to tokenize the pieces of text into an input. The tokens are converted into input IDs, and an attention mask is created - the both are then converted into PyTorch tensors. The tensors are fed into the pre-trained DistilBERT model to obtain the BERT embeddings. The mean pooling operation is applied to obtain a single embedding vector for the entire input. 
\begin{table}[htbp]
  \centering
  \begin{tabular}{c|p{6.5cm}}
    \toprule
    \textbf{Index} & \textbf{BERT Embeddings} \\
    \midrule
    0 & [-0.036851548, 0.10950465, 0.4777028...] \\
    1 & [-0.0088326335, 0.19127563, 0.47957394...] \\
    2 & [-0.17745975, 0.25770444, 0.23720546...] \\
    \bottomrule
  \end{tabular}
  \caption{BERT Embeddings}
\end{table}
Table 1 shows the embedding vectors for the first three song lyrics observations. After extracting the embedding vectors, we use them as input to predict the target variable 'year'. For doing so, a Support Vector Machine Regression (SVR) model with a linear kernel is initialized. When used for regression tasks, Support Vector Machines (SVMs) typically transfer input vectors into a high-dimensional feature space, where a linear regression is carried out. Because the link between the input vectors and the output values is frequently unclear and probably non-linear, this mapping is essential. In this high-dimensional space, SVR explicitly looks for a linear hyperplane that best fits the input vectors to the corresponding output values \cite{wauters2014support}. Our specific goal of this model's training is to translate the BERT embeddings to the target variable (e.g. forecasting a number connected to the lyrics). The SVR model learns the relationship between the BERT embeddings and the target variable.

\section{Experiment}

\subsection{Dataset}
For gaining insights and relationships in the world of music, we want to use a dataset which contains information as recent as 2022 scraped from Genius. In this place, music, poetry, and even novels can be uploaded and annotated (but largely songs). The dataset holds over 5 million lyric entries in many different languages. We access the data set via the 'Kaggle' website. For our project, we just want to focus on song lyrics written in English. Additionally, due to computation constraints, we want to use a randomly chosen sample for our models. The key variables in the dataset include the title, artist, lyrics, genre, year, and page views, with each playing a crucial role in understanding and analyzing the music data. 

In the dataset, the genre is labeled as 'tag' and includes pop, rock, rb, rap, misc, and country. Most of non-music pieces are labeled as 'misc', therefore we dropped all rows with the tag value 'misc'. Furthermore, we found out that the lowest value of the column year is 1. To have a more up to date analysis, we just kept songs in our final datasets which are older than 1960.

\subsection{Descriptive Analysis and Preprocessing}

As a first step, we started by collecting a distinct dataset from Genius. For our data descriptive analysis, we use a dataset with 50,000 observations. After some initial data inspection and exploration, an exploratory data analysis (EDA) was conducted. The EDA focused on understanding the composition of the dataset, including the genre distribution. We figured out, that the most common genres are pop and rap. We also analyzed the most frequent words in the different genres. We found out that the most frequent words in the genres pop, rock, country, and rb are similar where the most used word is 'love'. However, artists that are writing songs in the genre rap use other special words. For rap, the most frequent words include also 'bitch' and 'fuck'. Furthermore, we visualized the song lyric length and sought to find patterns that might affect model training. During our EDA we observed that songs lyrics with the genre rap have on average the highest word count and text length, and rock songs the lowest. However, we figured out that a few songs have a very low word count which we labeled as non-serious songs. That's why, we set a filter with lyrics word count above 100 words. Additionally, we performed a sentiment analysis, determining the emotional tone of song words. This analysis not only helps in revealing the emotional intricacies of individual songs but also investigates potential emotional links between genres. 
\begin{figure}[h]
\centering
\includegraphics[width=\linewidth]{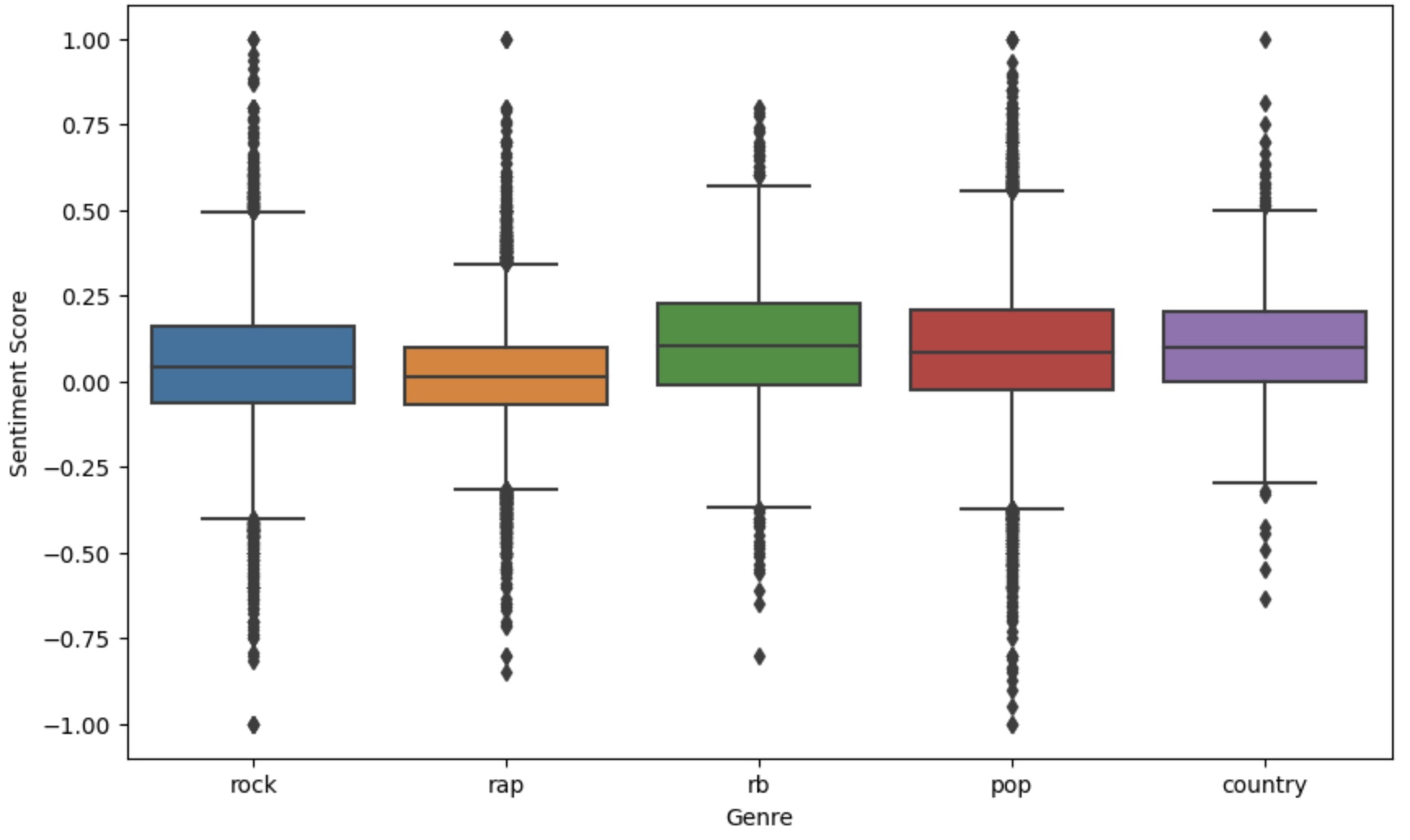}
\caption{Distribution of Sentiment Scores by Genre}
\end{figure}
Figure 3 shows the distribution of sentiment scores by genre. It is obvious to observe that the genre rap has the lowest sentiment score and therefore the highest negative sentiment. This is not surprising because as already mentioned above, the words used in rap lyrics are more negative than in the other genres. The most positive sentiment score can be observed in the genres rb and pop.
\begin{figure}[h]
\centering
\includegraphics[width=\linewidth]{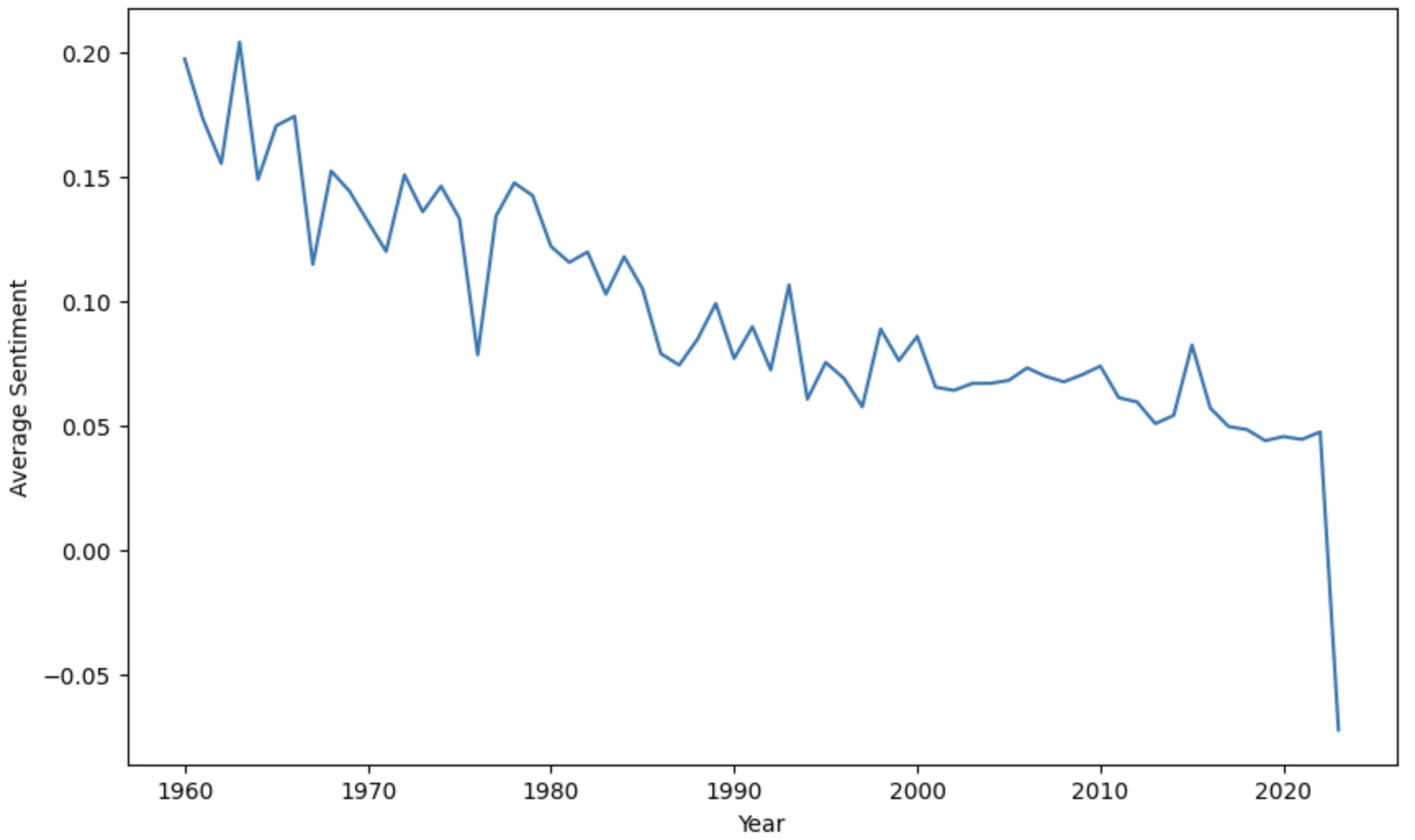}
\caption{Sentiment Trends over Years}
\end{figure}
Figure 4 analyzes the sentiment trend over the years from 1960 to 2022. We can see that the sentiment trend decreases over the years and song writers tend to a more neutral sentiment when writing their song lyrics. However, nowadays the sentiment score tends to be more negatively on average. A reason could be that more and more rap songs are released. However, a bias may exist due to a smaller sample size of songs in 2022. \\

Before implementing the BERT model, we decided to sample new datasets from our big Genius dataset for each of the three use cases. For predicting genres, we selected a dataset of 50,000 songs, focusing on entries with over 150 words to ensure substantial lyrical content and a minimum of 1,000 views to maintain a basic quality standard. This approach helped filter out non-song entries and outliers. Additionally, we balanced the dataset across various genres, adjusting sample sizes for more accurate and distinct genre predictions. To forecast the success of a song, we labeled songs with views under 100,000 as a fail, and songs with page views above 100,000 as success. We balance the dataset by 8,000 songs from both success classes to ensure an equal presentation of the training data. Finally, to predict the release year of a song, we extracted 300 songs for each year from 1960-2022 for having a good balance. Additionally, for the genre and year use case, we just consider songs with above 1,000 views because we wanted to focus on more popular songs. Another pre processing step is data cleaning for special characters in the song lyrics. The main structure of the lyrics is preserved and song metadata is frequently present between square brackets in the middle of the lyrics. This means that each entry probably has a lot of new line characters, which might be problematic when reading the data or feeding it to a model. \\

\subsection{Evaluation Metrics}
For our song lyrics classification task we use accuracy and F1-score as our evaluation metrics. Accuracy is the ratio of the correctly labeled genres and success/fail to the whole pool of songs \cite{baratloo2015part}. Equation (2) shows how accuracy is defined, where TP stands for true positive, TN for true negative, FP for false positive, FN is false negative, and TN stands for true negative. 
\begin{equation}
    \text{Accuracy} = \frac{\text{TP} + \text{TN}}{\text{TP} + \text{FP} + \text{FN} + \text{TN}}
\end{equation}
The F1-score can be interpreted as a harmonic mean of the precision and recall. Where precision is defined as $\frac{TP}{(TP+FP)}$ and recall as $\frac{TP}{(TP+FN)}$.
\begin{equation}
    \text{F1 Score} = \frac{2 \times (\text{Recall} \times \text{Precision})}{\text{Recall} + \text{Precision}}
\end{equation}
For predicting the release year, we implement the root mean squared error (RMSE) as our evaluation metrics. The RMSE is a measure of the average magnitude of errors between predicted values and actual values, which is shown in equation (4) \cite{chai2014root}.
\begin{equation}
    \text{RMSE} = \sqrt{\frac{1}{n} \sum_{i=1}^{n}e_i^2}
\end{equation}

\subsection{Model Results}
Genre classification presented a notable challenge, particularly in distinguishing between pop and rock lyrics, due to their shared thematic and stylistic nuances. We experimented with various methods, such as a two-step classification process and focus weight adjustments, in an effort to refine the model's accuracy between these genres. Ultimately, the solution lay in fine-tuning a standard model and hyperparameter optimization. This fine-tuning improved our model's discernment capabilities, leading to a genre classification accuracy of 65\% across five different classes, marking a substantial achievement within the domain of natural language processing for song lyrics. Furthermore, the genre classification for predicting rap has a F1-score of 0.84, which implies a very good prediction - not that surprising because of the unique words which are used in rap lyrics.
\begin{figure}[h]
\centering
\includegraphics[width=\linewidth]{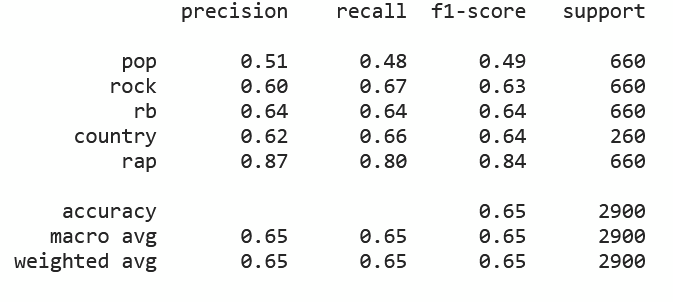}
\caption{Classification Report for Predicting Genre}
\end{figure}

Since it was pretty hard, to predict the exact page view for a specific song just based at song lyrics BERT embeddings, we decided to label songs by their success based on the views. The best performance could be observed at a one-split classification task, where we just classify the songs into success and fail. 
\begin{figure}[h]
\centering
\includegraphics[width=\linewidth]{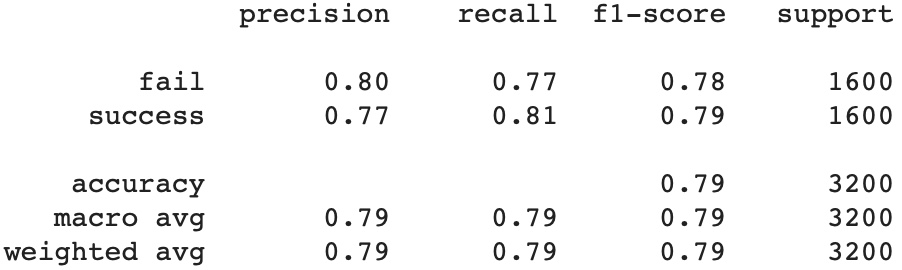}
\caption{Classification Report for Predicting Success}
\end{figure}
The classification report in figure 6 shows how good our model performs on our test data. With an overall accuracy of 79\% our model achieves good results in classifying songs as success or fail. 

Finally, we tried to predict the approximate release year of a song. For this, we trained several machine learning models with our extracted BERT embeddings. 
\begin{table}[htbp]
  \centering
  \begin{tabular}{lr}
    \toprule
    \textbf{Model} & \textbf{RMSE} \\
    \midrule
    Random Forest & 16.45 \\
    Support Vector Machines & 14.18 \\
    Linear Regression & 14.28 \\
    XGBoost & 15.26 \\
    K-Nearest Neighbors & 16.83 \\
    \bottomrule
  \end{tabular}
  \caption{RMSE Values for Different Models}
\end{table}
Table 2 shows the root mean squared errors for the different machine learning models. The table displays that all classic machine learning models performs similar based on the RMSE, with Support Vector Machines Regression Model performing the best with the lowest RMSE of 14.18. We also tried to fine tune a neural network based on our DistilBERT embeddings. However, due to computational restriction, we had difficulties to fine tune the neural network with its best hyperparamaters. Therefore, we ignore the results for our neural network in table 2.

\begin{figure}[h]
\centering
\includegraphics[width=\linewidth]{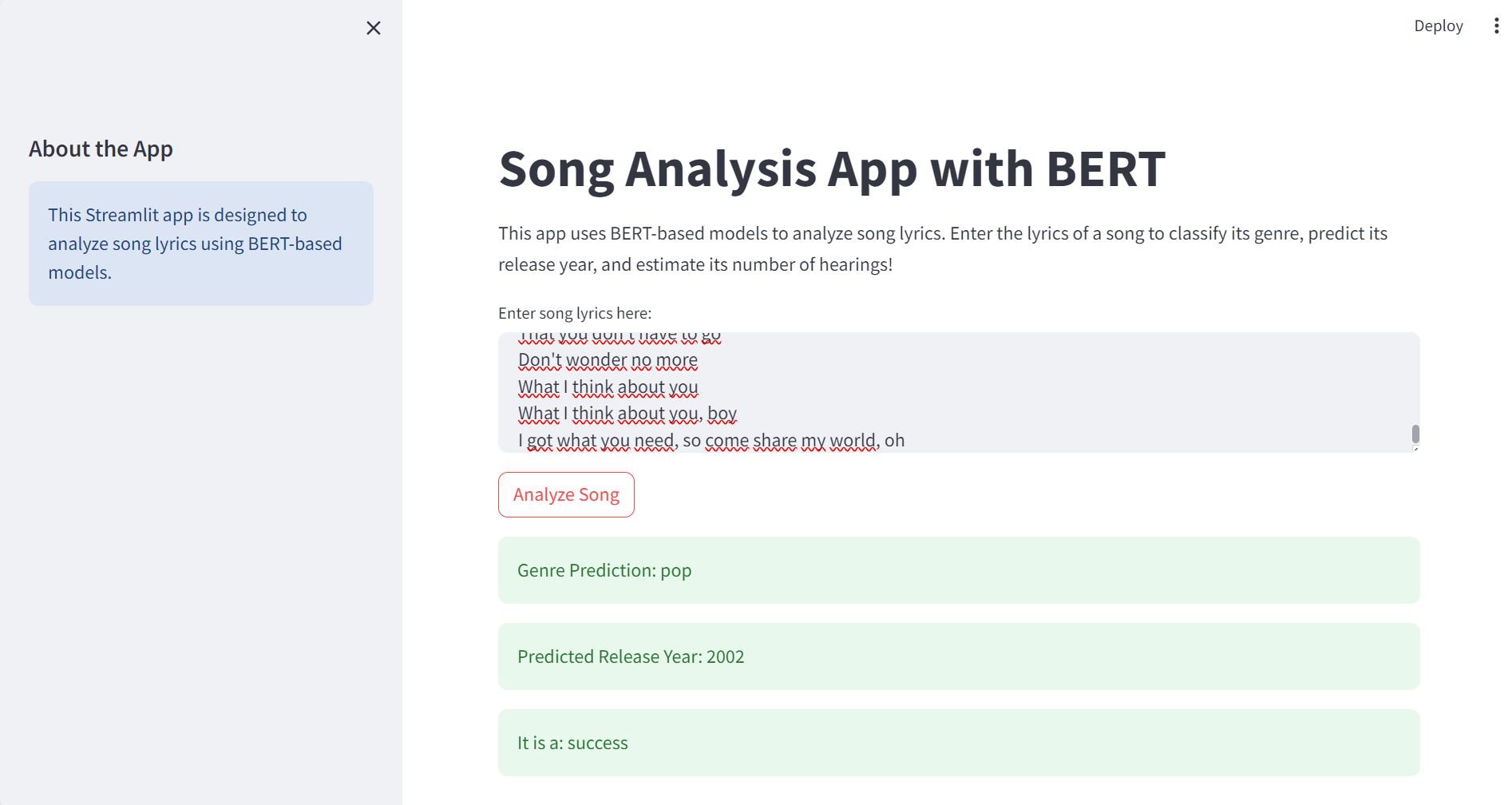}
\caption{Example of application prediction for Rihanna's song "If It’s Lovin’ That You Want" from 2005, illustrating the integrated output of our models.}
\label{fig:example-rihanna}
\end{figure}

In a final step we have integrated our fine tuned DistilBERT models into a practical app dashboard. This application utilizes our DistilBERT-based models to analyze input song lyrics and predict genre, potential success, and release year in an integrated fashion.
As an example, the dashboard has successfully identified the genre, predicted the success and approximated the release year for Rihanna's 2005 song "If It’s Lovin’ That You Want", as shown in figure 7. This not only demonstrates the apps's capabilities but also rounds up our fun and engaging approach to the classification of song lyrics.

\section{Conclusion}
In conclusion, this study has successfully leveraged the DistilBERT model to navigate the complexities of song lyric classification, tackling the various challenges of genre differentiation with a notable accuracy of 65\% and forecasting song success with a 79\% accuracy rate. The models' proficiency extends to estimating song release years, as evidenced by the lowest RMSE of 14.18 achieved by Support Vector Machines. By integrating these models into a user-friendly dashboard, we have provided an innovative tool that not only predicts genre, success, and release year with remarkable precision but also offers a playful yet insightful exploration into the emotional landscape of music lyrics. This fusion of advanced NLP techniques and practical application encapsulates the essence of our research, marking a significant stride in the understanding and classification of music only through its lyrical content.

\bibliographystyle{aaai}
\bibliography{reference}

\begin{thebibliography}{}

\bibitem[\protect\citeauthoryear{Adel \bgroup et al\mbox.\egroup }{2022}]{adel2022improving}
Adel, H.; Dahou, A.; Mabrouk, A.; Abd~Elaziz, M.; Kayed, M.; El-Henawy, I.~M.; Alshathri, S.; and Amin~Ali, A.
\newblock 2022.
\newblock Improving crisis events detection using distilbert with hunger games search algorithm.
\newblock {\em Mathematics} 10(3):447.

\bibitem[\protect\citeauthoryear{Baratloo \bgroup et al\mbox.\egroup }{2015}]{baratloo2015part}
Baratloo, A.; Hosseini, M.; Negida, A.; and El~Ashal, G.
\newblock 2015.
\newblock Part 1: simple definition and calculation of accuracy, sensitivity and specificity.

\bibitem[\protect\citeauthoryear{Chai and Draxler}{2014}]{chai2014root}
Chai, T., and Draxler, R.~R.
\newblock 2014.
\newblock Root mean square error (rmse) or mean absolute error (mae)?--arguments against avoiding rmse in the literature.
\newblock {\em Geoscientific model development} 7(3):1247--1250.

\bibitem[\protect\citeauthoryear{Chen \bgroup et al\mbox.\egroup }{2023}]{chen2023adaptive}
Chen, Y.; Li, Z.; Ouyang, W.; and Lepech, M.
\newblock 2023.
\newblock Adaptive hierarchical spatiotemporal network for traffic forecasting.
\newblock In {\em 2023 IEEE 19th International Conference on Automation Science and Engineering (CASE)},  1--6.
\newblock IEEE.

\bibitem[\protect\citeauthoryear{Devlin \bgroup et al\mbox.\egroup }{2018}]{devlin2018bert}
Devlin, J.; Chang, M.-W.; Lee, K.; and Toutanova, K.
\newblock 2018.
\newblock Bert: Pre-training of deep bidirectional transformers for language understanding.
\newblock {\em arXiv preprint arXiv:1810.04805}.

\bibitem[\protect\citeauthoryear{Guo \bgroup et al\mbox.\egroup }{2024}]{guo2024online}
Guo, P.; Jin, P.; Li, Z.; Bai, L.; and Zhang, Y.
\newblock 2024.
\newblock Online test-time adaptation of spatial-temporal traffic flow forecasting.
\newblock {\em arXiv preprint arXiv:2401.04148}.

\bibitem[\protect\citeauthoryear{Jiang \bgroup et al\mbox.\egroup }{2023}]{jiang2023unified}
Jiang, M.; Wang, A.; Li, Z.; and Tsung, F.
\newblock 2023.
\newblock A unified probabilistic framework for spatiotemporal passenger crowdedness inference within urban rail transit network.
\newblock In {\em 2023 IEEE 19th International Conference on Automation Science and Engineering (CASE)},  1--8.
\newblock IEEE.

\bibitem[\protect\citeauthoryear{Jiang \bgroup et al\mbox.\egroup }{2024}]{jiang2024x}
Jiang, H.; Li, Z.; Wei, H.; Xiong, X.; Ruan, J.; Lu, J.; Mao, H.; and Zhao, R.
\newblock 2024.
\newblock X-light: Cross-city traffic signal control using transformer on transformer as meta multi-agent reinforcement learner.
\newblock In {\em IJCAI 2024: The 33rd International Joint Conference on Artificial Intelligence}.

\bibitem[\protect\citeauthoryear{Kong \bgroup et al\mbox.\egroup }{2024}]{kong2024tptu}
Kong, Y.; Ruan, J.; Chen, Y.; Zhang, B.; Bao, T.; Shi, S.; Du, G.; Hu, X.; Mao, H.; Li, Z.; et~al.
\newblock 2024.
\newblock Tptu-v2: Boosting task planning and tool usage of large language model-based agents in real-world systems.
\newblock In {\em ICLR 2024: The Twelfth International Conference on Learning Representations Workshop on LLM Agents}.

\bibitem[\protect\citeauthoryear{Li \bgroup et al\mbox.\egroup }{2020a}]{li2020tensor}
Li, Z.; Sergin, N.~D.; Yan, H.; Zhang, C.; and Tsung, F.
\newblock 2020a.
\newblock Tensor completion for weakly-dependent data on graph for metro passenger flow prediction.
\newblock In {\em AAAI 2024:}, volume~34,  4804--4810.
\newblock AAAI Technical Track: Machine Learning.

\bibitem[\protect\citeauthoryear{Li \bgroup et al\mbox.\egroup }{2020b}]{li2020long}
Li, Z.; Yan, H.; Zhang, C.; and Tsung, F.
\newblock 2020b.
\newblock Long-short term spatiotemporal tensor prediction for passenger flow profile.
\newblock {\em IEEE Robotics and Automation Letters} 5(4):5010--5017.

\bibitem[\protect\citeauthoryear{Li \bgroup et al\mbox.\egroup }{2022a}]{li2022profile}
Li, Z.; Yan, H.; Tsung, F.; and Zhang, K.
\newblock 2022a.
\newblock Profile decomposition based hybrid transfer learning for cold-start data anomaly detection.
\newblock {\em ACM Transactions on Knowledge Discovery from Data (TKDD)} 16(6):1--28.

\bibitem[\protect\citeauthoryear{Li \bgroup et al\mbox.\egroup }{2022b}]{li2022individualized}
Li, Z.; Yan, H.; Zhang, C.; and Tsung, F.
\newblock 2022b.
\newblock Individualized passenger travel pattern multi-clustering based on graph regularized tensor latent dirichlet allocation.
\newblock {\em Data Mining and Knowledge Discovery} 36(4):1247--1278.

\bibitem[\protect\citeauthoryear{Li \bgroup et al\mbox.\egroup }{2023a}]{li2023critical}
Li, D.; Li, Z.; Li, Z.; Bai, L.; Gong, Q.; Sun, L.; Ketter, W.; and Zhao, R.
\newblock 2023a.
\newblock A critical perceptual pre-trained model for complex trajectory recovery.
\newblock In {\em SIGSPATIAL 2023: 31st ACM SIGSPATIAL Conference: International Workshop on Searching and Mining Large Collections of Geospatial Data (GeoSearch '23)},  1--9.

\bibitem[\protect\citeauthoryear{Li \bgroup et al\mbox.\egroup }{2023b}]{li2023visiontraj}
Li, Z.; Li, Z.; Hu, X.; Du, G.; Nie, Y.; Zhu, F.; Bai, L.; and Zhao, R.
\newblock 2023b.
\newblock Visiontraj: A noise-robust trajectory recovery framework based on large-scale camera network.
\newblock {\em arXiv preprint arXiv:2312.06428}.

\bibitem[\protect\citeauthoryear{Li \bgroup et al\mbox.\egroup }{2023c}]{li2023tensor}
Li, Z.; Yan, H.; Zhang, C.; Ketter, W.; and Tsung, F.
\newblock 2023c.
\newblock Tensor dirichlet process multinomial mixture model with graphs for passenger trajectory clustering.
\newblock In {\em SIGSPATIAL 2023: Proceedings of the 6th ACM SIGSPATIAL International Workshop on AI for Geographic Knowledge Discovery},  121--128.

\bibitem[\protect\citeauthoryear{Li \bgroup et al\mbox.\egroup }{2023d}]{li2023choose}
Li, Z.; Yan, H.; Zhang, C.; Sun, L.; Ketter, W.; and Tsung, F.
\newblock 2023d.
\newblock Choose a table: Tensor dirichlet process multinomial mixture model with graphs for passenger trajectory clustering.
\newblock {\em arXiv preprint arXiv:2310.20224}.

\bibitem[\protect\citeauthoryear{Li \bgroup et al\mbox.\egroup }{2023e}]{li2023tensorb}
Li, Z.; Yan, H.; Zhang, C.; Wang, A.; Ketter, W.; Sun, L.; and Tsung, F.
\newblock 2023e.
\newblock Tensor dirichlet process multinomial mixture model for passenger trajectory clustering.
\newblock {\em arXiv preprint arXiv:2306.13794}.

\bibitem[\protect\citeauthoryear{Li \bgroup et al\mbox.\egroup }{2024}]{li2024non}
Li, Z.; Nie, Y.; Li, Z.; Bai, L.; Lv, Y.; and Zhao, R.
\newblock 2024.
\newblock Non-neighbors also matter to kriging: A new contrastive-prototypical learning.
\newblock In {\em AISTATS 2024: Proceedings of The 27th International Conference on Artificial Intelligence and Statistics}.

\bibitem[\protect\citeauthoryear{Lin \bgroup et al\mbox.\egroup }{2023}]{lin2023dynamic}
Lin, J.; Li, Z.; Li, Z.; Bai, L.; Rui, Z.; and Zhang, C.
\newblock 2023.
\newblock Dynamic causal graph convolutional network for traffic prediction.
\newblock In {\em 2023 IEEE 19th International Conference on Automation Science and Engineering (CASE)},  1--8.
\newblock IEEE.

\bibitem[\protect\citeauthoryear{Liu \bgroup et al\mbox.\egroup }{2023}]{liu2023relation}
Liu, K.; Tang, S.; Li, Z.; Li, Z.; Bai, L.; Zhu, F.; and Zhao, R.
\newblock 2023.
\newblock Relation-aware distribution representation network for person clustering with multiple modalities.
\newblock {\em IEEE Transactions on Multimedia}.

\bibitem[\protect\citeauthoryear{Liu \bgroup et al\mbox.\egroup }{2024a}]{liu2024timecma}
Liu, C.; Xu, Q.; Miao, H.; Yang, S.; Zhang, L.; Long, C.; Li, Z.; and Zhao, R.
\newblock 2024a.
\newblock Timecma: Towards llm-empowered time series forecasting via cross-modality alignment.
\newblock {\em arXiv preprint arXiv:2406.01638}.

\bibitem[\protect\citeauthoryear{Liu \bgroup et al\mbox.\egroup }{2024b}]{liu2024spatial}
Liu, C.; Yang, S.; Xu, Q.; Li, Z.; Long, C.; Li, Z.; and Zhao, R.
\newblock 2024b.
\newblock Spatial-temporal large language model for traffic prediction.
\newblock In {\em IEEE MDM 2024: The 25th IEEE International Conference on Mobile Data Management}.

\bibitem[\protect\citeauthoryear{Lu \bgroup et al\mbox.\egroup }{2024}]{lu2024dualight}
Lu, J.; Ruan, J.; Jiang, H.; Li, Z.; Mao, H.; and Zhao, R.
\newblock 2024.
\newblock Dualight: Enhancing traffic signal control by leveraging scenario-specific and scenario-shared knowledge.
\newblock In {\em AAMAS 2024: The 23rd International Conference on Autonomous Agents and Multiagent Systems}.

\bibitem[\protect\citeauthoryear{Mao \bgroup et al\mbox.\egroup }{2022a}]{mao2022transformer}
Mao, H.; Zhao, R.; Li, Z.; Chen, H.; Hao, J.; Chen, Y.; Li, D.; Zhang, J.; and Xiao, Z.
\newblock 2022a.
\newblock Transformer in transformer as backbone for deep reinforcement learning.
\newblock {\em arXiv preprint arXiv:2212.14538}.

\bibitem[\protect\citeauthoryear{Mao \bgroup et al\mbox.\egroup }{2022b}]{mao2022jointly}
Mao, Z.; Li, Z.; Li, D.; Bai, L.; and Zhao, R.
\newblock 2022b.
\newblock Jointly contrastive representation learning on road network and trajectory.
\newblock In {\em CIKM 2022: Proceedings of the 31st ACM International Conference on Information \& Knowledge Management},  1501--1510.

\bibitem[\protect\citeauthoryear{Mao \bgroup et al\mbox.\egroup }{2024}]{mao2024pdit}
Mao, H.; Zhao, R.; Li, Z.; Xu, Z.; Chen, H.; Chen, Y.; Zhang, B.; Xiao, Z.; Zhang, J.; and Yin, J.
\newblock 2024.
\newblock Pdit: Interleaving perception and decision-making transformers for deep reinforcement learning.
\newblock In {\em AAMAS 2024: The 23rd International Conference on Autonomous Agents and Multiagent Systems}.

\bibitem[\protect\citeauthoryear{Rajendran, Pillai, and Daneshfar}{2022}]{rajendran2022lybert}
Rajendran, R.~V.; Pillai, A.~S.; and Daneshfar, F.
\newblock 2022.
\newblock Lybert: Multi-class classification of lyrics using bidirectional encoder representations from transformers (bert).

\bibitem[\protect\citeauthoryear{Ruan \bgroup et al\mbox.\egroup }{2023a}]{ruan2023privacy}
Ruan, H.; Gong, Q.; Chen, Y.; Chen, J.; Li, Z.; and Su, X.
\newblock 2023a.
\newblock A privacy-preserving heart rate prediction system for drivers in connected vehicles.
\newblock In {\em Proceedings of the 21st Annual International Conference on Mobile Systems, Applications and Services},  557--558.

\bibitem[\protect\citeauthoryear{Ruan \bgroup et al\mbox.\egroup }{2023b}]{ruan2023tptu}
Ruan, J.; Chen, Y.; Zhang, B.; Xu, Z.; Bao, T.; Du, G.; Shi, S.; Mao, H.; Li, Z.; Zeng, X.; et~al.
\newblock 2023b.
\newblock Tptu: Task planning and tool usage of large language model-based ai agents.
\newblock In {\em NeurIPS 2023: 37th Conference on Neural Information Processing Systems (NeurIPS 2023) - Workshop on Foundation Models for Decision Making}.

\bibitem[\protect\citeauthoryear{Ruan \bgroup et al\mbox.\egroup }{2024}]{ruan2024coslight}
Ruan, J.; Li, Z.; Wei, H.; Jiang, H.; Lu, J.; Xiong, X.; Mao, H.; and Zhao, R.
\newblock 2024.
\newblock Coslight: Co-optimizing collaborator selection and decision-making to enhance traffic signal control.
\newblock In {\em SIGKDD 2024: Proceedings of the 30th ACM SIGKDD Conference on Knowledge Discovery and Data Mining}.

\bibitem[\protect\citeauthoryear{Sanh \bgroup et al\mbox.\egroup }{2019}]{sanh2019distilbert}
Sanh, V.; Debut, L.; Chaumond, J.; and Wolf, T.
\newblock 2019.
\newblock Distilbert, a distilled version of bert: smaller, faster, cheaper and lighter.
\newblock {\em arXiv preprint arXiv:1910.01108}.

\bibitem[\protect\citeauthoryear{Sergin~Nurretin \bgroup et al\mbox.\egroup }{2024}]{sergin2024low}
Sergin~Nurretin, D.; Hu, J.; Li, Z.; Zhang, C.; Tsung, F.; and Yan, H.
\newblock 2024.
\newblock Low-rank robust subspace tensor clustering for metro passenger flow modeling.
\newblock {\em INFORMS Journal on Data Science}.

\bibitem[\protect\citeauthoryear{Tsung and Li}{2020}]{tsung2020discussion}
Tsung, F., and Li, Z.
\newblock 2020.
\newblock Discussion of “a novel approach to the analysis of spatial and functional data over complex domains”.
\newblock {\em Quality Engineering} 32(2):193--196.

\bibitem[\protect\citeauthoryear{Wang \bgroup et al\mbox.\egroup }{2023}]{wang2023correlated}
Wang, L.; Bai, L.; Li, Z.; Zhao, R.; and Tsung, F.
\newblock 2023.
\newblock Correlated time series self-supervised representation learning via spatiotemporal bootstrapping.
\newblock In {\em 2023 IEEE 19th International Conference on Automation Science and Engineering (CASE)},  1--7.
\newblock IEEE.

\bibitem[\protect\citeauthoryear{Wauters and Vanhoucke}{2014}]{wauters2014support}
Wauters, M., and Vanhoucke, M.
\newblock 2014.
\newblock Support vector machine regression for project control forecasting.
\newblock {\em Automation in Construction} 47:92--106.

\bibitem[\protect\citeauthoryear{Xu \bgroup et al\mbox.\egroup }{2023}]{xu2023kits}
Xu, Q.; Long, C.; Li, Z.; Ruan, S.; Zhao, R.; and Li, Z.
\newblock 2023.
\newblock Kits: Inductive spatio-temporal kriging with increment training strategy.
\newblock {\em arXiv preprint arXiv:2311.02565}.

\bibitem[\protect\citeauthoryear{Yan \bgroup et al\mbox.\egroup }{2024}]{yan2024sparse}
Yan, H.; Li, Z.; Zhao, X.; and Hu, J.
\newblock 2024.
\newblock Sparse decomposition methods for spatio-temporal anomaly detection.
\newblock In {\em Multimodal and Tensor Data Analytics for Industrial Systems Improvement}. Springer.
\newblock  185--206.

\bibitem[\protect\citeauthoryear{Ye \bgroup et al\mbox.\egroup }{2024}]{ye2024survey}
Ye, J.; Zhang, W.; Yi, K.; Yu, Y.; Li, Z.; Li, J.; and Tsung, F.
\newblock 2024.
\newblock A survey of time series foundation models: Generalizing time series representation with large language mode.
\newblock {\em arXiv preprint arXiv:2405.02358}.

\bibitem[\protect\citeauthoryear{Zhang \bgroup et al\mbox.\egroup }{2024a}]{zhang2024controlling}
Zhang, B.; Mao, H.; Ruan, J.; Wen, Y.; Li, Y.; Zhang, S.; Xu, Z.; Li, D.; Li, Z.; Zhao, R.; et~al.
\newblock 2024a.
\newblock Controlling large language model-based agents for large-scale decision-making: An actor-critic approach.
\newblock In {\em ICLR 2024: The Twelfth International Conference on Learning Representations Workshop on LLM Agents}.

\bibitem[\protect\citeauthoryear{Zhang \bgroup et al\mbox.\egroup }{2024b}]{zhang2024dualtime}
Zhang, W.; Ye, J.; Li, Z.; Li, J.; and Tsung, F.
\newblock 2024b.
\newblock Dualtime: A dual-adapter multimodal language model for time series representation.
\newblock {\em arXiv e-prints}  arXiv--2406.

\bibitem[\protect\citeauthoryear{Ziyue}{2021}]{ziyue2021tensor}
Ziyue, L.
\newblock 2021.
\newblock Tensor topic models with graphs and applications on individualized travel patterns.
\newblock In {\em ICDE 2021: 2021 IEEE 37th International Conference on Data Engineering (ICDE)},  2756--2761.
\newblock IEEE.

\end{thebibliography}

%\section{References}

%Genius Dataset \url{https://www.kaggle.com/datasets/carlosgdcj/genius-song-lyrics-with-language-information}. \\

%Khrulenko, I. (2020). Pop, rap or heavy metal - lyrics classifier. \textit{Kaggle}. \url{https://www.kaggle.com/code/ivankhrulenko/pop-rap-or-heavy-metal-lyrics-classifier}. \\

\end{document}